\documentclass[11pt,a4paper]{article}
\usepackage[hyperref]{emnlp2020}
\usepackage{times}
\usepackage{latexsym}
\usepackage[noend]{algpseudocode}
\usepackage{url}
\usepackage{todonotes}
\usepackage{graphicx}
\usepackage{comment}
\usepackage{amsmath}
\usepackage{bm}
\usepackage{soul}
\usepackage[normalem]{ulem}

\usepackage{microtype}

\aclfinalcopy 

\usepackage[flushleft]{threeparttable}
\usepackage{todonotes}
\usepackage{enumitem}
\usepackage{algorithm}
\usepackage{algpseudocode}
\usepackage{booktabs}
\usepackage{subcaption}
\usepackage[nointegrals]{wasysym}
\usepackage{tikz}
\usepackage{tabu}
\usepackage{tabularx}
\usepackage{float}
\usepackage{listings}
\usepackage{color}
\usepackage{multirow}
\usepackage{framed}
\usepackage{pifont}
\usepackage{dsfont}
\usepackage{xcolor}
\usepackage{amsmath,amsthm}
\usepackage{amssymb}
\usepackage{graphicx}
\usepackage{textcomp}
\usepackage{tcolorbox}
\usepackage{soul}
\usepackage{todonotes}

\definecolor{DarkRed}{RGB}{130,25,0}

\newcommand{\type}[1]{{\tt#1}}

\newcommand{\tbd}{\textrm{TimeBank-Dense}}

\newcommand{\ib}{\textsc{I2b2-Temporal}}

\newcommand{\ignore}[1]{}
\definecolor{mypurple}{RGB}{121,55,196}
\definecolor{myblue}{RGB}{60,177,245}
\definecolor{myorange}{RGB}{243,206,84}
\definecolor{mygreen}{RGB}{41,222,35}

\newcounter{exctr}

\newcounter{eventCtr}

\newcounter{timexCtr}

\newcommand{\temprel}[1]{\MakeUppercase{\textit{#1}}}
\newcommand{\event}[1]{\textbf{#1}}

\title{
Domain Knowledge Empowered Structured Neural Net for End-to-End Event Temporal Relation Extraction
}
\author{Rujun Han$^{1,2}$ ~ Yichao Zhou$^{3}$ ~ Nanyun Peng$^{1,2,3}$\\
$^1$Information Sciences Institute, University of Southern California \\
$^2$Department of Computer Science, University of Southern California \\
$^3$Department of Computer Science, University of California, Los Angeles \\
{\tt rujunhan@isi.edu; yz@cs.ucla.edu; violetpeng@cs.ucla.edu}}

\date{}

\begin{document}
\maketitle

\begin{abstract}

Extracting event temporal relations is a critical task for information extraction and plays an important role in natural language understanding. Prior systems leverage deep learning and pre-trained language models to improve the performance of the task. However, these systems often suffer from two shortcomings: 1) when performing maximum a posteriori (MAP) inference based on neural models, previous systems only used structured knowledge that is assumed to be absolutely correct, i.e., \textit{hard constraints}; 2) biased predictions on dominant temporal relations when training with a limited amount of data. To address these issues, we propose a framework that enhances deep neural network with \textit{distributional constraints} constructed by probabilistic domain knowledge. We solve the constrained inference problem via Lagrangian Relaxation and apply it to end-to-end event temporal relation extraction tasks. Experimental results show our framework is able to improve the baseline neural network models with strong statistical significance on two widely used datasets in news and clinical domains.
\end{abstract}
\section{Introduction}
\label{sec:intro}

\begin{figure}[t]
    \centering
    \includegraphics[angle=-90, clip, trim=2cm 0cm 2cm 0cm, width=0.48\textwidth]{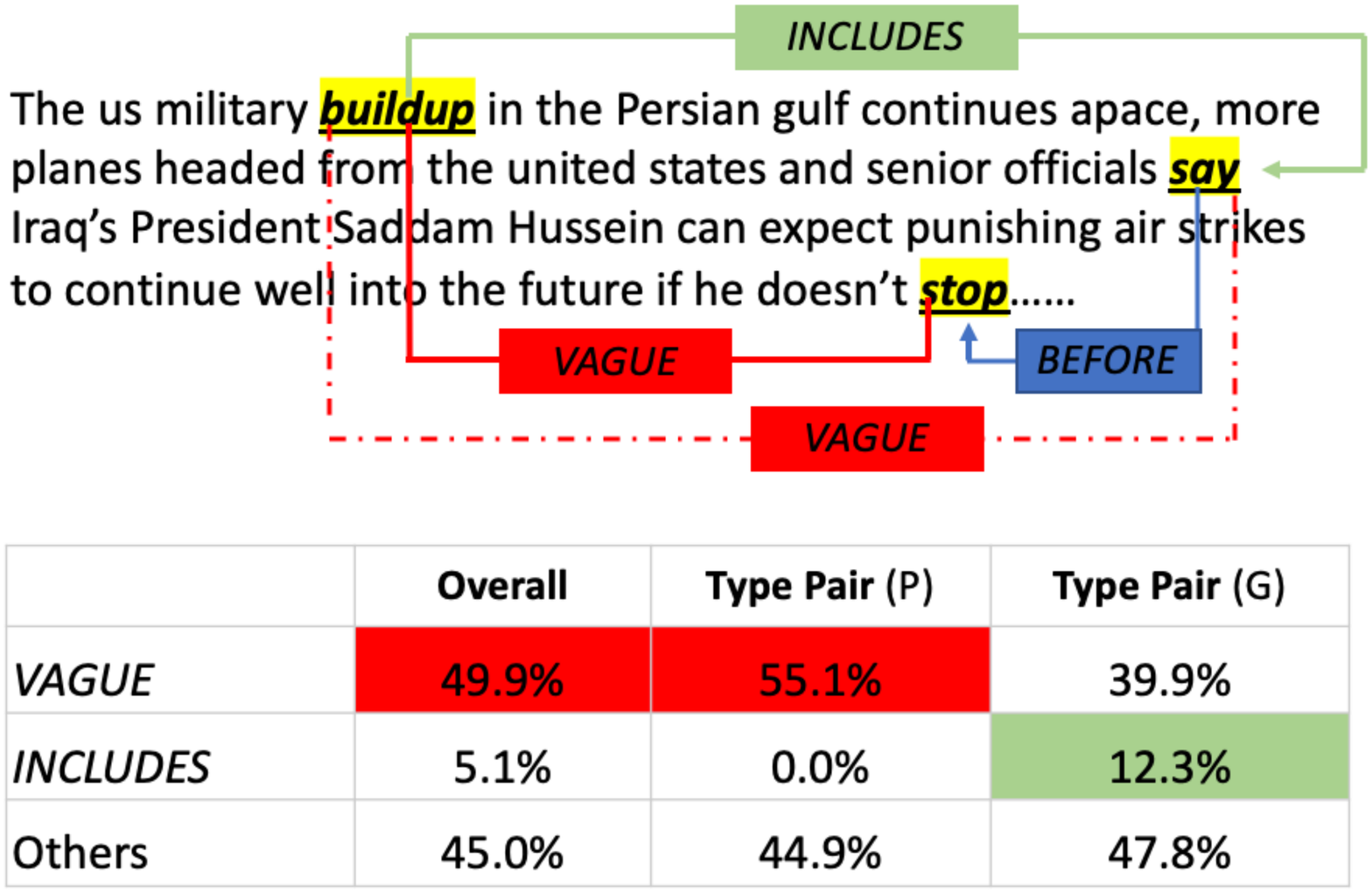}
    \caption{An example of the event temporal ordering task. Text input is taken from the news dataset in our experiments. Solid lines / arrows between two highlighted events show their gold temporal relations, e.g. \event{say} \temprel{before} \event{stop} and \event{buildup} \temprel{includes} \event{say},  and the dash line shows a wrong prediction, i.e., the \temprel{vague} relation between \event{buildup} and \event{say}. In the table, Column Overall shows the relation distribution over the entire training corpus; Column Type Pair (P) shows the \textit{predicted} relation distribution condition on the event pairs having types \type{occurrence} and \type{reporting} (such as \event{buildup} and \event{say}); Column Type Pair (G) shows the \textit{gold} relation distribution condition on event pairs having the same types. Biased predictions of \temprel{vague} relation between \event{buildup} and \event{say} can be partially corrected by using the gold event type-relation statistics in Column Type Pair (G).}
    \vspace{-0.5cm}
    \label{fig:temRel-ex}
\end{figure}

Extracting event temporal relations from raw text data has attracted surging attention in the NLP research community in recent years as it is a fundamental task for commonsense reasoning and natural language understanding. It facilitates various downstream applications, such as forecasting social events and tracking patients' medical history. Figure~\ref{fig:temRel-ex} shows an example of this task where an event extractor first needs to identify events (\event{buildup}, \event{say} and \event{stop}) in the input and then a relation classifier predicts all pairwise relations among them, resulting in a temporal ordering as illustrated in the figure. For example, \event{say} is \temprel{before} \event{stop}; \event{buildup} \temprel{includes} \event{say}; the temporal ordering between \event{buildup} and \event{stop} cannot be decided from the context, so the relation should be \temprel{vague}.

Predicting event temporal relations is inherently challenging as it requires the system to understand each event's beginning and end times. However, these time anchors are often hard to specify within a complicated context, even for humans. As a result, there is usually a large amount of \temprel{vague} pairs (nearly 50\% in the table of Figure~\ref{fig:temRel-ex}) in an expert-annotated dataset, resulting in heavily class-imbalanced datasets. Moreover, expert annotations are often time-consuming to gather, so the sizes of existing datasets are relatively small. To cope with the class-imbalance problem and the small dataset issues, recent research efforts adopt \textit{hard constraint}-enhanced deep learning methods and leverage pre-trained language models \cite{ning-etal-2018-cogcomptime, han-etal-2019-joint} and are able to establish reasonable baselines for the task.

The \textit{hard-constraints} used in the SOTA systems can only be constructed when they are nearly 100\% correct and hence make the knowledge adoption restrictive. Temporal relation transitivity is a frequently used \textit{hard constraint} that requires \textit{if} \event{A} \temprel{before} \event{B} and \event{B} \temprel{before} \event{C}, it \textit{must be} that \event{A} \temprel{before} \event{C}. However, constraints are usually not deterministic in real-world applications. For example, a clinical \type{treatment} and \type{test} are more \textit{likely} to happen \temprel{after} a medical \type{problem}, but not \textit{always}. Such probabilistic constraints cannot be encoded with the \textit{hard-constraints} as in the previous systems.

Furthermore, deep neural models have biased predictions on dominant classes, which is particularly concerning given the small and biased datasets in event temporal extraction. For example, in Figure~\ref{fig:temRel-ex}, an event pair \event{headed} and \event{say} (with relation \temprel{includes}) is incorrectly predicted as \temprel{vague} (Column Type Pair (P)) by our baseline neural model, partially due to dominant percentage of \temprel{vague} label (Column Overall), and partially due to the complexity of the context. Using the domain knowledge that \event{headed} and \event{say} have event types of \type{occurrence} and \type{reporting}, respectively, we can find a new label probability distribution (Type Pair (G)) for this pair. The probability mass allocated to \temprel{vague} would decrease by 10\% and increase by 7.2\% for \temprel{includes}, which significantly increases the chance for a correct label prediction.

We propose to improve deep structured neural networks by incorporating domain knowledge such as corpus statistics in the model inference, and by solving the constrained inference problem using Lagrangian Relaxation. This framework allows us to benefit from the strong contextual understanding of pre-trained language models while optimizing model outputs based on probabilistic structured knowledge that previous deep models fail to consider. Experimental results demonstrate the effectiveness of this framework.

We summarize our contributions below: 

\begin{itemize}
\itemsep0em
    \item We formulate the incorporation of probabilistic knowledge as a constrained inference problem and use it to optimize the outcomes from strong neural models.
    \item Novel applications of Lagrangian Relaxation on end-to-end temporal relation extraction task with event-type and relation constraints.
    \item Our framework significantly outperforms baseline systems without knowledge adoption and achieves new SOTA results on two datasets in news and clinical domains.
    \vspace{-0.4cm}
\end{itemize}

\begin{figure*}[t]
    \centering
    \includegraphics[angle=-90, clip, trim=5.5cm 0cm 5.6cm 0cm, width=\textwidth]{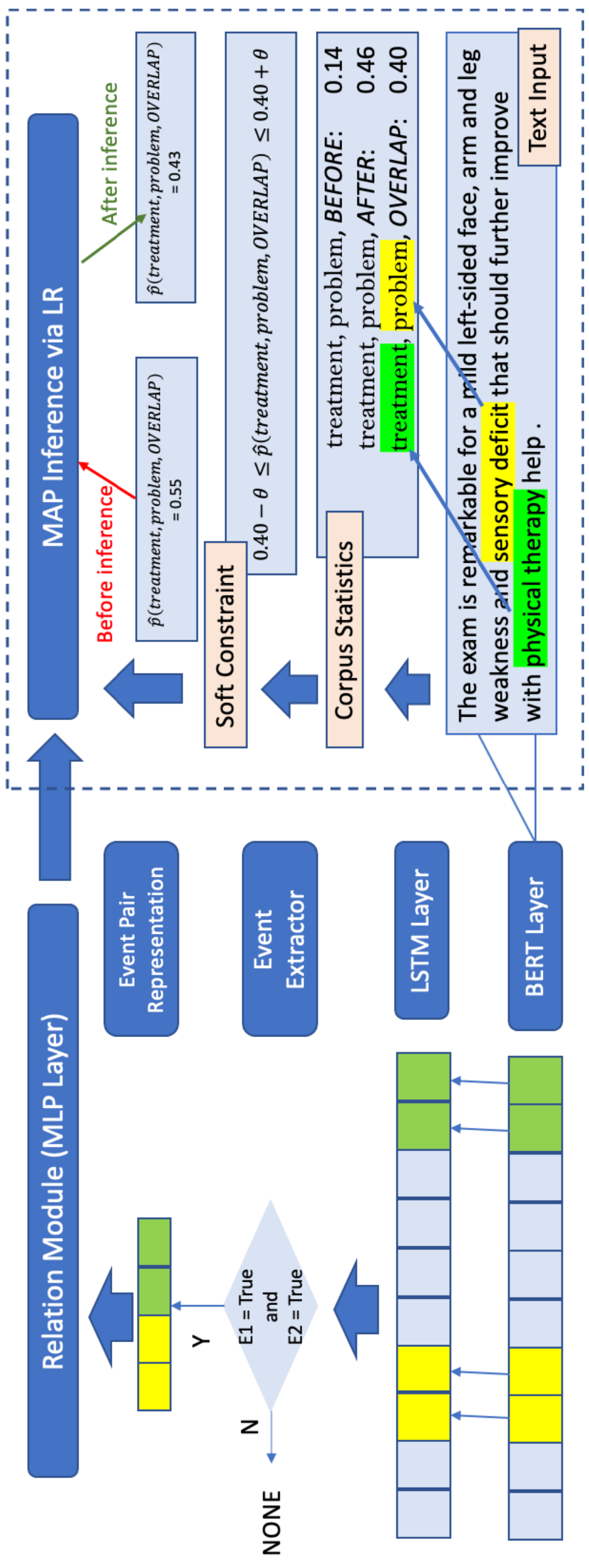}
    \caption{An overview of the proposed framework. The left column shows the end-to-end event temporal relation extraction workflow. The right column (in the dashed box) illustrates how we propose to enhance the end-to-end extraction system. The final MAP inference contains two components: scores from the relation module and \textit{distributional constraints} constructed using domain knowledge and corpus statistics. The text input is a real example taken from the {\ib} dataset. The MAP inference is able to push the predicted probability of the event type-relation triplet closer to the ground-truth (corpus statistics).}
    \label{fig:framework}
    \vspace{-0.4cm}
\end{figure*}

\section{Problem Formulation}
\label{sec:problem}

The problem we focus on is end-to-end event temporal relation extraction, which takes a raw text as input, first identifies all events, and then classifies temporal relations for all predicted event pairs. The left column of Figure~\ref{fig:framework} shows an example. An end-to-end system is practical in a real-world setting where events are not annotated in the input and challenging because temporal relations are harder to predict after noise is introduced during event extraction. 
\section{Method}
\label{sec:method}
In this section, we first describe the details of our deep neural networks for an end-to-end event temporal relation extraction system, then show how to formulate \textit{domain-knowledge} between event types and relations as \textit{distributional constraints} in Integer Linear Programming (ILP), and finally apply Lagrangian Relaxation to solve the constrained inference problem. Our base model is trained end-to-end with cross-entropy loss and multitask learning to obtain relation scores. We need to perform an additional inference step in order to incorporate \textit{domain-knowledge} as \textit{distributional constraints}.

\subsection{End-to-end Event Relation Extraction}
As illustrated in the left column in Figure~\ref{fig:framework}, our end-to-end model shares a similar work-flow as the pipeline model in \citet{han-etal-2019-joint}, where multi-task learning with a shared feature extractor is used to train the pipeline model. Let $\mathcal{E}$, $\mathcal{E}\mathcal{E}$ and $\mathcal{R}$ denote event, candidate event pairs and the feasible relations, respectively, in an input instance $\boldsymbol{x}^n$, where $n$ is the instance index. The combined training loss is $\mathcal{L} = c_{\mathcal{E}}\mathcal{L}_{\mathcal{E}} + \mathcal{L}_\mathcal{R}$,
where $\mathcal{L}_{\mathcal{E}}$ and $\mathcal{L}_\mathcal{R}$ are the losses for the event extractor and the relation module, respectively, and 
$c_{\mathcal{E}}$ is a hyper-parameter balancing the two losses.

\paragraph{Feature Encoder.}
Input instances are first sent to pre-trained language models such as BERT \cite{BERT2018} and RoBERTa \cite{ROBERTA-19}, then to a Bi-LSTM layer as in previous event temporal relation extraction work \cite{han-etal-2019-deep}. 
 
Encoded features will be used as inputs to the event extractor and the relation module below.
\vspace{-0.1cm}
\paragraph{Event Extractor.} The event extractor first predicts scores over event classes for each input token and then detects event spans based on these scores. If an event has over more than one tokens, its beginning and ending vectors are concatenated as the final event representation. The event score is defined as the predicted probability distribution over event classes. Pairs predicted to include non-events are automatically labeled as \temprel{none}, whereas valid candidate event pairs are fed into the relation module to obtain their relation scores.
\paragraph{Relation Module.} 
The relation module's input is a pair of events, which share the same encoded features as the event extractor. We simply concatenate them before feeding them into the relation module to produce relation scores $S(y^r_{i,j},\boldsymbol{x}^n)$, which are computed using the Softmax function
where $y^r_{i,j}$ is a binary indicator of whether an event pair $(i,j) \in \mathcal{E}\mathcal{E}$ has relation $r \in \mathcal{R}$.

\subsection{Constrained Inference for Knowledge Incorporation} 
As shown in Figure ~\ref{fig:framework}, once the relation scores are computed via the relation module, a MAP inference is performed to incorporate \textit{distributional constraints} so that the structured knowledge can be used to adjust neural baseline model scores and optimize the final model outputs. We formulate our MAP inference with \textit{distributional constraints} as an LR problem and solve it with an iterative algorithm.

Next, we explain the details of each component in our MAP inference.
\vspace{-0.2cm}
\subsubsection{Distributional constraints}
Much of the \textit{domain-knowledge} required for real-world problems are probabilistic in nature. In the task of event relation extraction, \textit{domain-knowledge} can be the prior probability of a specific event-pair's occurrence acquired from large corpora or knowledge base \cite{ning-etal-2018-improving}; \textit{domain-knowledge} can also be event-property and relation distribution obtained using corpus statistics, as we study in this work. Previous work mostly leverage \textit{hard constraints} for inference \citep{yoshikawa2009jointly, NingFeRo17, leeuwenberg2017structured, P18-1212, han-etal-2019-deep, han-etal-2019-joint}, where constraints such as transitivity and event-relation consistency are assumed to be absolutely correct. As we discuss in Section \ref{sec:intro}, \textit{hard constraints} are rigid and thus cannot be used to model probabilistic  \textit{domain-knowledge}.

The right column in Figure~\ref{fig:framework} illustrates how our work leverages corpus statistics to construct \textit{distributional constraints}. Let $P$ be a set of event properties such as clinical types (e.g. \type{treatment} or \type{problem}). 

For the pair $(P^m, P^n)$ and the triplet $(P^m, P^n, r)$, where $P^n, P^m \in P$ and $r \in \mathcal{R}$, we can retrieve their counts in the training corpus as
{\small
\[
C(P^m, P^n, r) = \sum_{i,j \in \mathcal{E}\mathcal{E}} c(P_i = P^m; P_j = P^n; r_{i,j} = r)
\]
}
\vspace{-0.2cm}
and 
{\small
\[
C(P^m, P^n) = \sum_{i,j \in \mathcal{E}\mathcal{E}} c(P_i = P^m; P_j = P^n).
\]
} Let $t=(P^m, P^n, r)$. The prior triplet probability can thus be defined as
\begin{align}
p^*_t = & \frac{C(P^m, P^n, r)}{C(P^m, P^n)}. \nonumber
\end{align}
Let $\hat{p}_t$ denote the predicted triplet probability, \textit{distributional constraints} require that,
\begin{align} \label{eq:constraint}
p^*_t - \theta \le \hat{p}_t \le p^*_t + \theta
\end{align}
where $\theta$ is the tolerance margin between the prior and predicted probabilities.

\subsubsection{Integer Linear Programming with Distributional Constraints}
\label{sec:ilp}
We formulate our MAP inference as an ILP problem. Let $\mathcal{T}$ be a set of triplets whose predicted probabilities need to satisfy Equation~\ref{eq:constraint}. We can define our full ILP as
{
\begin{align} \label{eq:infObj-full}
\mathcal{L} = & \sum_{(i,j) \in \mathcal{E}\mathcal{E}} \sum_{r \in \mathcal{R}} y^r_{i,j} S(y^r_{i,j},\boldsymbol{x})
\end{align}
{\small
\[
\textbf{s.t.      } p^*_t - \theta \le \hat{p}_t \le p^*_t + \theta, \text{   }\forall t \in \mathcal{T} \text{,    and}
\]
\[
y^r_{i,j} \in \{0, 1\} \text{  , } \sum_{r \in \mathcal{R}} y^r_{i,j} = 1,
\]
}
}where $S(y^r_{i,j},\boldsymbol{x}), \forall r \in \mathcal{R}$ is the scoring function obtained from the relation module. For $t = (P^m, P^n, r)$, we have $
\hat{p}_t = \frac{\sum^{\mathcal{E}\mathcal{E}}_{(i:P^m, j:P^n)}y^r_{i,j}} {\sum^{\mathcal{E}\mathcal{E}}_{(i:P^m, j:P^n)} \sum_{r'}^\mathcal{R}y^{r'}_{i,j}}
$.
The output of the MAP inference, $\bf{\hat{y}}$, is a collection of optimal label assignments for all relation candidates in an input instance $\boldsymbol{x}^n$. $\sum_{r \in\mathcal{R}} y^r_{i,j} = 1$ ensures that each event pair gets one label assignment and this is the only \textit{hard constraint} we use. 

To improve computational efficiency, we apply the heuristic to optimize only the equality constraints $p^*_t = \hat{p_t}$, $\forall t \in \mathcal{T}$. Our optimization algorithm terminates when $|p^*_t - \hat{p_t}| \le \theta$. This heuristic has been shown to work efficiently without hurting inference performance \citep{meng-etal-2019-target}. For each triplet $t$, its equality constraint can be rewritten as
\begin{align} \label{eq:substitute}
F(t) = (1 - p_t^*) \sum^{\mathcal{E}\mathcal{E}}_{(i:P^m, j:P^n)} y^r_{i,j}, & \\- p_t^*\sum^{\mathcal{E}\mathcal{E}}_{(i: P^m, j:P^n,} \sum^{\mathcal{R}}_{r' \ne r)} y^{r'}_{i,j} & = 0. \nonumber
\end{align} 
The goal is to maximize the objective function defined by Eq.~\eqref{eq:infObj-full} while satisfying the equality constraints.
\begin{algorithm}
\small
\caption{Gradient Ascent for LR}
\begin{algorithmic}[1]
\Procedure{}{} 
    \For{$t \in \mathcal{T}$}
    \State $\lambda_t^0 = 0$ 
    \EndFor
    \State $k = 0$
    \While{$k < K $} \Comment{$K$: max iteration}
    \State $\hat{y}^{k+1} \leftarrow \arg \max \mathcal{L}(\bm{\lambda}^k)$
        \For{$t \in \mathcal{T}$}
            \State $\Delta_t = p^*_t - \hat{p}_t$
            \If{$|\Delta_t| > \theta$} 
            \State $\lambda_t^{k+1} = \lambda_t^k + \alpha\Delta_t$
            \EndIf
        \EndFor
        \If{$\Delta_t \le \theta, \forall t$} 
        \State \textbf{break}
        \EndIf
        \State $k = k + 1$
        \State $\alpha = \gamma\alpha$ \Comment{$\gamma$: decay rate}
    \EndWhile
\EndProcedure
\end{algorithmic}
\end{algorithm}
\vspace{-0.5cm}
\subsubsection{Lagrangian Relaxation}
\label{sec:lr}
Solving Eq.~\eqref{eq:infObj-full} is NP-hard. Thus, we reformulate it as a Lagrangian Relaxation problem by introducing Lagrangian multipliers $\lambda_t$ for each \textit{distributional constraint}. Lagrangian Relaxation has been applied in a variety NLP tasks, as described by \citet{rush-collins-2011-exact, Rush-Collins-tutorial-12} and \citet{zhao-etal-2017-men}. 

The Lagrangian Relaxation problem can be written as
\begin{align} \label{eq:infObj-LR}
\mathcal{L}(\mathbf{y}, \bm{\lambda}) = & \sum_{(i,j) \in \mathcal{E}\mathcal{E}} \sum_{r \in \mathcal{R}} y^r_{i,j} S(y^r_{i,j},\boldsymbol{x}) + \sum_{t \in \mathcal{T}}\lambda_t F(t).
\end{align}

Initialize $\lambda_t = 0$. Eq.~\eqref{eq:infObj-LR} can be solved with the following iterative algorithm (Algorithm 1). 
\vspace{-0.2cm}
\begin{enumerate}
\itemsep0em
    \item[1.] At each iteration $k$, obtain the best relation assignments per MAP inference, $
    \hat{y}^k = \arg \max \mathcal{L}(\mathbf{y}, \bm{\lambda})
    $
    \item[2.] Update the Lagrangian multiplier in order to bring the predicted probability closer to the prior. Specifically, for each $t \in \mathcal{T}$,
    \begin{itemize}
    \itemsep0em
        \item If $|p^*_t - \hat{p}_t| \le \theta$, $\lambda_t^{k+1} = \lambda_t^{k}$
        \item Otherwise, $\lambda_t^{k+1} = \lambda_t^k + \alpha(p_t^* - \hat{p_t})$
    \end{itemize}
\end{enumerate}
$\alpha$ is the step size. We are solving a min-max problem: the first step chooses the maximum likelihood assignments by fixing $\bm{\lambda}$; the second step searches for $\bm{\lambda}$ values that minimize the objective function.

\section{Constrained Inference Implementation} 
\label{sec:imp}
This section explains how to construct our \textit{distributional constraints} and the implementation details for inference with LR.

\begin{table}[t]
\centering
\begin{tabular}{@{ }l@{ }c@{ }c@{ }}
\toprule
Constraint Triplets & Count  & $\%$  \\ \hline \hline
\type{occurrence}, \type{occurrence}, * & 124 & 19.7 \\
\type{occurrence}, \type{reporting}, * & 50 & 7.9 \\ 
\type{occurrence}, \type{action}, * & 44   & 7.0 \\ 
\type{reporting}, \type{occurrence}, * & 41  & 6.5 \\
\type{action}, \type{occurrence}, * & 40 & 6.4 \\
\type{action}, \type{action}, * & 20 & 3.2 \\
\hline\hline
\type{reporting}, \type{reporting}, * & 18 & 2.9 \\ 
\type{action}, \type{reporting}, * & 18   & 2.9 \\ 
\type{reporting}, \type{action}, * & 17  & 2.7 \\
\bottomrule
\end{tabular}
\caption{{\tbd}: triplet prediction count and percentage in the development set (sample size = 629).}
\label{tab:tbd-triplets}
\vspace{-0.3cm}
\end{table}

\begin{table*}[htbp!]
\centering
\small
\begin{tabular}{lcccc|ccccc}
\toprule
& \multicolumn{4}{c|}{\textbf{\tbd}} & \multicolumn{5}{c}{\textbf{2012 i2b2 Challenge} (\ib)}\\
\cmidrule(lr){2-5}\cmidrule(lr){6-10}
& \multicolumn{1}{c}{Event} & \multicolumn{3}{c|}{Relation}& \multicolumn{2}{c}{Event}& \multicolumn{3}{c}{Relation (TempEval Metrics)}\\
\cmidrule(lr){2-2}\cmidrule(lr){3-5}\cmidrule(lr){6-7}\cmidrule(lr){8-10} \\
& F$_1$ & R & P & F$_1$ & Span F$_1$  & Type Accuracy & R &  P & F$_1$\\
\cmidrule(lr){1-1}\cmidrule(lr){2-2}\cmidrule(lr){3-3}\cmidrule(lr){4-4}\cmidrule(lr){5-5}\cmidrule(lr){6-6}\cmidrule(lr){7-7}\cmidrule(lr){8-8}\cmidrule(lr){9-9}\cmidrule(lr){10-10}
{\em Feature-based Benchmark} & {87.4} & {43.8} & {35.7} & {39.4} & { \textbf{90.1}} & {86.0} & {37.8} & {51.8} & {43.0} \\
{\citet{han-etal-2019-joint}} & {\textbf{90.9}} & {52.6} & {46.5} & {49.4} & {-} & {-} & {73.4} & {76.3} & {74.8} \\
End-to-end Baseline& 90.3 & 51.5 & 45.9 & 48.5 & 87.8 & \textbf{87.8} & 73.3 & 79.9 & 76.5\\
End-to-end + Inference & 90.3 & \textbf{53.4} & \textbf{47.9} & \textbf{50.5} & 87.8 & \textbf{87.8} & \textbf{74.0} & \textbf{80.8} & \textbf{77.3} \\
\bottomrule
\end{tabular}
\caption{Overall experiment results: per MacNemar's test, the improvements against the end-to-end baseline models by adding inference with \textit{distributional constraints} are both statistically significant for {\tbd} (p-value $<0.005$) and {\ib} (p-value $<0.0005$). For {\ib}, our end-to-end system is optimized for the F$_1$ score of the gold pairs.}
\label{tab:result}
\vspace{-.5cm}
\end{table*}

\subsection{Distributional Constraint Selection}
The selection of \textit{distributional constraints} is crucial for our algorithm. If the probability of an event-type and relation triplet is unstable across different splits of data, we may over-correct the predicted probability. We use the following search algorithm with heuristic rules to ensure constraint stability.

\subsubsection{\tbd}
For {\tbd}, we first sort candidate constraints by their corresponding values of $C(P^m, P^n) = \sum_{\hat{r}\in\mathcal{R}}C(P^m, P^n, \hat{r})$. We list $C(P^m, P^n)$ with the largest prediction numbers and their percentages in the development set in Table~\ref{tab:tbd-triplets}.

Next, we set $3\%$ as our threshold to include constraints for our main experimental results. We found this number to work relatively well for both {\tbd} and {\ib}. We will show the impact of relaxing this threshold in the discussion section. In Table~\ref{tab:tbd-triplets}, the constraints in the bottom block are filtered out. Moreover, Eq.~\ref{eq:substitute} implies that a constraint defined on one triplet $(P^m, P^n, r)$ has impact on all $(P^m, P^n, r')$ for $r' \in \mathcal{R} \setminus r$. In other words, decreasing $\hat{p}_{(P^m, P^n, r)}$ is equivalent to increasing $\hat{p}_{(P^m, P^n, r')}$ and vice versa. Thus, we heuristically pick $(P^m, P^n,\temprel{vague})$ as our default constraint triplets. 

Finally, we adopt a greedy search rule to select the final set of constraints. We start with the top constraint triplet in Table~\ref{tab:tbd-triplets} and then keep adding the next one as long as it doesn't hurt the grid search\footnote{Recall that our LR algorithm in Section~\ref{sec:lr} has three hyper-parameters: initial step size $\alpha$, decay rate $\gamma$, and tolerance $\theta$. We perform a grid search on the development set and use the best hyper-parameters on the test set.} F$_1$ score on the development set. Eventually, four constraints triplets are selected, and they can be found in Table~\ref{tab:tbd-abl}.

\subsubsection{\ib}
Similar to {\tbd}, we use the $3\%$ threshold to select candidate constraints. However, it is computationally expensive to use the greedy search rule above by conducting grid search as the number of constraints that pass this threshold is large (15 of them), development set sample size is more than 3 times of {\tbd}, and a large transformer is used for modeling,  Therefore, we incorporate another two heuristic rules to directly select constraints,
\begin{enumerate}
\setlength\itemsep{0em}
\item We randomly split the train data into five subsets of equal size $\{s_1, s_2, s_3, s_4, s_5\}$. For triplet $t$ to be selected, we must have $\frac{1}{5}\sum^5_{k=1} |p_{t, s_k} - p^*_t| < 0.001$. 
\item $|\hat{p}_t - p^*_t| > 0.1$, where $\hat{p}_t$ is the predicted probability of $t$ on the development set.
\end{enumerate}
The first rule ensures that a constraint triplet is stable over a randomly split of data; the second one ensures that the probability gaps between the predicted and gold are large so that we will not over-correct them. Eventually, four constraints satisfy these rules, and they can be found in Table~\ref{tab:i2b2-abl}, and we run only one final grid search for these constraints. 

\subsection{Inference}
The ILP component in Sec.~\ref{sec:ilp} is implemented using an off-the-shelf solver provided by Gurobi optimizer. Hyper-parameters choices can be found in Table~\ref{tab:hyper} in the Appendix.

\section{Experimental Setup}
\label{sec:setup}

This section describes the two event temporal relation datasets used in this paper and then explains the evaluation metrics.

\subsection{Data}
\label{sec:data}
\paragraph{\tbd.} Temporal relation corpora such as TimeBank \citep{PustejovskyX2003} and RED \citep{O'Gorman2016} consist of expert annotations of news articles. The common issue of these corpora is missing annotations. Collecting densely annotated temporal relation corpora with all events and relations fully annotated is a challenging task as annotators could easily overlook some facts \citep{Bethard:2007:TTI:1304608.1306306, P14-2082, ChambersTBS2014, NingFeRo17}. 

The {\tbd} dataset mitigates this issue by forcing annotators to examine all pairs of events within the same or neighboring sentences, and this dataset has been widely evaluated on this task \cite{ChambersTBS2014, NingFeRo17, cheng2017classifying, meng2018context}. Temporal relations consist of \temprel{before}, \temprel{after}, \temprel{includes}, \temprel{Included}, \temprel{simultaneous}, and \temprel{vague}. Moreover, each event has several properties, e.g., \textbf{type}, \textbf{tense}, and \textbf{polarity}. Event types include \type{occurrence}, \type{action}, \type{reporting}, \type{state}, etc. Event pairs that are more than 2 sentences away are not annotated. 

\paragraph{\ib.} In the clinical domain, one of the earliest event temporal datasets was provided in the 2012 Informatics for Integrating Biology and the Bedside (i2b2) Challenge on NLP for Clinical Records \cite{rumshisky-etal-2013-eval}. Clinical events are categorized into 6 types: \type{treatment}, \type{problem}, \type{test}, \type{clinical-dept}, \type{occurrence}, and \type{evidential}. The final data used in the challenge contains three temporal relations: \temprel{before}, \temprel{after}, and \temprel{overlap}. The 2012 i2b2 challenge also had an end-to-end track, which we use as our feature-based system baseline. To mimic the input structure of \tbd, we only consider event pairs that are within 3 consecutive sentences. Overall, 13\% of the long-distance relations are excluded.\footnote{Over 80\% of these long-distance pairs are event co-reference, i.e., simply predicting them as \temprel{overlap} will achieve high performance.}

\subsection{Evaluation Metrics}
\label{sec:evaluation}

To be consistent with previous work, we adopt two different evaluation metrics. For \tbd, we use standard \textbf{micro-average} scores that are also used in the baseline system \cite{han-etal-2019-joint}. Since the end-to-end system can predict the gold pair as \temprel{none}, we follow the convention of IE tasks and exclude them from the evaluation. For \ib, we adopt the \textbf{TempEval} evaluation metrics used in the 2012 i2b2 challenge. These evaluation metrics differ from the standard F$_1$ in a way that it computes the graph closure for both gold and predictions labels. Since {\ib} contains roughly six times more missing annotations than the gold pairs, we only evaluate the performance of the gold pairs.

Both datasets contain three types of entities: events, time expressions, and document time. In this work, we focus on \textbf{event-event} relations and exclude all other relations from the evaluation.

\subsection{Baselines}
\label{sec:baselines}
\paragraph{Feature-based Systems.}
We use CAEVO\footnote{\url{ https://www.usna.edu/Users/cs/nchamber/caevo/}} \citep{ChambersTBS2014}, a hybrid system of rules and linguistic feature-based MaxEnt classifier, as our feature-based benchmark for {\tbd}. Model implementation and performance are both provided by \citet{han-etal-2019-joint}. As for {\ib}, we retrieve the predictions from the top end-to-end system provided by \citet{xu-etal-2013-end-to-end} and report the performance according to the evaluation metrics specified in Section~\ref{sec:evaluation}.
 
\paragraph{Neural Model Baselines.}
We use the end-to-end systems described by \citet{han-etal-2019-joint} as our neural network model benchmarks (Row 2 of Table~\ref{tab:result}). For {\tbd}, the best global structured model's performance is reported by \citet{han-etal-2019-joint}. For {\ib}, we re-implement the pipeline joint model. \footnote{\url{https://github.com/PlusLabNLP/JointEventTempRel}} Note that this end-to-end model only predicts whether each token is an event as well as each pair of token's relation. Event spans are not predicted, so head-tokens are used to represent events; event types are also not predicted. Therefore, we do not report Span F$_1$ and Type Accuracy in this benchmark.

\paragraph{End-to-end Baseline.}
For the \textit{\tbd} dataset, we use the pipeline joint (local) model with no global constraints as presented by \citet{han-etal-2019-joint}. In contrast to the aforementioned neural baseline provided in the same paper, this end-to-end model does not use any inference techniques. Hence, it serves as a fair baseline for our method (with inference). For \textit{\tbd}, we build our framework based on this model\footnote{Code and data for {\tbd} are published here: \url{https://github.com/rujunhan/EMNLP-2020}}.

For the \textit{\ib} dataset to be more comparable with the 2012 i2b2 challenge, we augment the event extractor illustrated in Figure~\ref{fig:framework} by allowing event type predictions; that is, for each input token, we not only predict whether it is an event or not, but also predict its event type. We follow the convention in the IE field by adding a ``BIO'' label to each token in the data. For example, the two tokens in ``physical therapy'' in Figure~\ref{fig:framework} will be labeled as B-\type{treatment} and I-\type{treatment}, respectively. To be consistent with the partial match method used in the 2012 i2b2 challenge, the event span detector looks for token predictions that start with either ``B-'' or ``I-'' and ensures that all tokens predicted within the same event span have only one event type. 

RoBERTa-large is used as the base model, and cross-entropy loss is used to train the model. We fine-tune the base model and conduct a grid search on the random hold-out set to pick the best hyper-parameters such as $c_{\mathcal{E}}$ in the multitask learning loss and the weight, $w_{\mathcal{E}_{pos}}$ for positive event types (i.e. B- and I-). The best hyper-parameter choices can be found in Table~\ref{tab:hyper} in the Appendix.
\section{Results and Analysis}
\vspace{-0.2cm}
\label{sec:result}
Table~\ref{tab:result} contains our main results. We discuss model performances on {\tbd} and {\ib} in this section.

\subsection{\tbd} All neural models outperform the feature-based system by more than 10\% per relation F$_1$ score. Our structured model outperforms the previous SOTA systems with \textit{hard constraints} and \textbf{joint event and relation training} by 1.1\%. Compared with the end-to-end baseline model with no constraints, our system achieves 2\% absolute improvement, which is statistically significant with a $p$-value $< 0.005$ per MacNemar's test. This is strong evidence that leveraging Lagrangian Relaxation to incorporate domain knowledge can be extremely beneficial even for strong neural network models.

The ablation study in Table~\ref{tab:tbd-abl} shows how \textit{distributional constraints} work and the constraints' individual contributions. The predicted probability gaps shrink by 0.15, 0.24, and 0.13 respectively for the three constraints chosen, while providing 0.91\%, 0.65\%, and 0.44\% improvements to the final F$_1$ score for relation extraction. We also show the breakdown of the performance for each relation class in Table~\ref{tab:tbd-breakdown}. The overall F$_1$ improvement is mainly driven by the recall scores in the positive relation classes (\temprel{before}, \temprel{after}, and \temprel{includes}) that have much smaller sample size than \temprel{vague}. These results are consistent with the ablation study in Table~\ref{tab:tbd-abl}, where the end-to-end baseline model over-predicts on \temprel{vague}, and the LR algorithm corrects it by assigning less confident predictions on \temprel{vague} to positive and minority classes according to their relation scores.

\subsection{\ib}
All neural models outperform the feature-based system by more than 30\% per relation F$_1$ score. Our structured model with \textit{distributional constraints} outperforms the neural pipeline joint models of \citet{han-etal-2019-joint} by 2.5\% per absolute scale. Compared with our end-to-end baseline model, our system achieves 0.77\% absolute improvement on F$_1$ measure, which is statistically significant with a $p$-value $< 0.0005$ per MacNemar's test. This result also shows that adding inference with \textit{distributional constraints} can be helpful for strong neural baseline models.

Table~\ref{tab:i2b2-abl} in the Appendix Section~\ref{sec:i2b2-results} shows how \textit{distributional constraints} work and their individual contributions. Predicted probability gaps shrink by 0.17, 0.16, 0.11, and 0.14, respectively, for the four constraints chosen, providing 0.19\%, 0.25\%, 0.22\%, and 0.12\% improvements to the final F$_1$ scores for relation extraction. We also have the breakdown performance for each relation class in Table~\ref{tab:i2b2-breakdown}. The performance gain is caused mostly by the increase of recall scores in \temprel{before} and \temprel{after}. This is consistent with the results in Table~\ref{tab:i2b2-abl} where the model over-predicts on the \temprel{overlap} class, possibly because of label imbalance. Inference is able to partially correct this mistake by leveraging \textit{distributional constraints} constructed with event type and relation corpus statistics. 

\begin{table}[t]
\centering
\small
\begin{tabular}{lcc}
\toprule
Constraint Triplets & Prob. Gap  & F$_1$  \\ \hline \hline
\type{occur.}, \type{occur.}, \temprel{vague} & -0.15 & +0.91\% \\
\type{occur.}, \type{reporting}, \temprel{vague} & -0.24 & +0.65\% \\ 
\type{action}, \type{occur.}, \temprel{vague} & -0.13   & +0.44\% \\ 
\type{reporting}, \type{occur.}, \temprel{vague}$^*$ & 0.0 & 0\% \\ 
\hline\hline
Combined F1 Improvement & & 2.0\% \\
\bottomrule
\end{tabular}
\caption{{\tbd} ablation study: gap shrinkage of predicted probability and F$_1$ contribution per constraint. $^*$ is selected per Sec.~\ref{sec:imp}, but the probability gap is smaller than the tolerance in the test set, hence no impact to the F$_1$ score.}
\label{tab:tbd-abl}
\vspace{-0.3cm}
\end{table}

\begin{table}[t]
    \small 
 	\centering
 	\setlength{\tabcolsep}{0.5em}
 	\begin{tabular}{|@{ }c@{ }|c@{ }|c@{ }|c@{ }|c@{ }|c@{ }|c@{ }|} \hline
 	&\multicolumn{3}{|@{ }c@{ }|}{\textbf{End-to-end Baseline}}
 	& \multicolumn{3}{|@{ }c@{ }|}{\textbf{End-to-end Inference}}\\ \cline{2-7}
 	& \textbf{P}& \textbf{R} & \textbf{F$_1$}&\textbf{P}& \textbf{R} & \textbf{F$_1$}\\ \hline
 	 \textbf{B}& 59.0& 46.9& 52.3 & 58.6& 55.7& 57.1\\
 	 \textbf{A} & 69.3 & 45.3 & 54.8 & 67.8& 51.5& 58.5\\
 	 \textbf{I}&-&-&-&8.3&1.8&2.9\\
 	 \textbf{II}&-&-&-&-&-&-\\
 	 \textbf{S}&-&-&-&-&-&-\\
 	 \textbf{V}&45.1&55.0&49.5 & 47.6& 51.4&49.4\\\hline \hline
 	 \textbf{Avg}&51.5&  45.9& 48.5 & 53.4 & 47.9 & 50.5\\ \hline
 	\end{tabular}
   	\caption{Model performance breakdown for {\tbd}. ``-'' indicates no predictions were made for that particular label, probably due to the small size of the training sample. \small{\temprel{Before} (\textbf{B}), \temprel{After} (\textbf{A}), \temprel{Includes} (\textbf{I}), \temprel{Is\_Included} (\textbf{II}), \temprel{Simultaneous} (\textbf{S}), \temprel{vague} (\textbf{V})}}
   	\vspace{-0.5cm}
   	\label{tab:tbd-breakdown}
 \end{table}

\subsection{Qualitative Error Analysis}
We can use the errors made by our structured neural model on {\tbd} to guide potential directions for future research. There are 26 errors made by the structured model that are correctly predicted by the baseline model. In Table~\ref{tab:error-analysis}, we show the error breakdown by constraints. Our method works by leveraging corpus statistics to correct borderline errors made by the baseline model; however, when the baseline model makes borderline correct predictions, the inference could mistakenly change them to the wrong labels. This situation can happen when the context is complicated or when the event time interval is confusing.

For the constraint (\type{occur.}, \type{occur.}, \temprel{vague}), nearly all errors are cross-sentence event pairs with long context information. In \textbf{ex.1}, the gold relation between \event{responded} and \event{use} is \temprel{vague} because of the negation of \event{use}, but one could also argue that if \event{use} were to happen, \event{responded} is \temprel{before} \event{use}. This inherent annotation confusion can cause the baseline model to predict \temprel{vague} marginally over \temprel{before}. When informed by the constraint statistics that \event{vague} is over-predicted, the inference algorithm revises the baseline prediction to \temprel{before}. Similarly, in \textbf{ex.2} and \textbf{ex.3}, one could make strong cases that both the relations between \event{delving} and \event{acknowledged}, and \event{opposed} and \event{expansion} are \temprel{before} rather than \temprel{vague} from the context. This annotation ambiguity can contribute to the errors made by the proposed method.

Our analysis shows that besides the necessity to create high-quality data for event temporal relation extraction, it could be useful to incorporate additional information such as discourse relation (particularly for (\type{occur.}, \type{occur.}, \temprel{vague})) and other prior knowledge on event properties to resolve the ambiguity in event temporal reasoning.

\begin{table}[t]
    \small 
 	\centering
 	\begin{tabular}{l} \hline
 	\toprule
 	\type{occurrence}, \type{occurrence}, \temprel{vague} (57.7\%) \\\hline
 	\textbf{ex.1} In a bit of television diplomacy, Iraq's deputy \\ foreign minister \hl{responded} from Baghdad in less than \\ one hour, saying Washington would break international \\ law by attacking without UN approval. The United States \\ is not authorized to \hl{use} force before going to the council. \\\hline\hline
 	\type{occurrence}, \type{reporting}, \temprel{vague} (26.9\%) \\\hline
 	\textbf{ex.2} A new Essex County task force began \hl{delving} \\ Thursday into the slayings of 14 black women over the \\ last five years in the Newark area, as law-enforcement \\officials \hl{acknowledged} that they needed to work harder... \\\hline\hline
 	\type{action}, \type{occurrence}, \temprel{vague} (15.4\%) \\\hline
 	\textbf{ex.3} The Russian leadership has staunchly \hl{opposed} \\ the western alliance's \hl{expansion} into Eastern Europe. \\\bottomrule
 	\end{tabular}
   	\caption{Error examples and breakdown by constraints.}
   	\vspace{-0.5cm}
   	\label{tab:error-analysis}
 \end{table}
\section{Discussion}
\subsection{Constraint Selection}
\begin{figure}[t]
    \centering
    \includegraphics[angle=-90, clip, trim=2.5cm 1cm 3.5cm 1cm, width=0.49\textwidth]{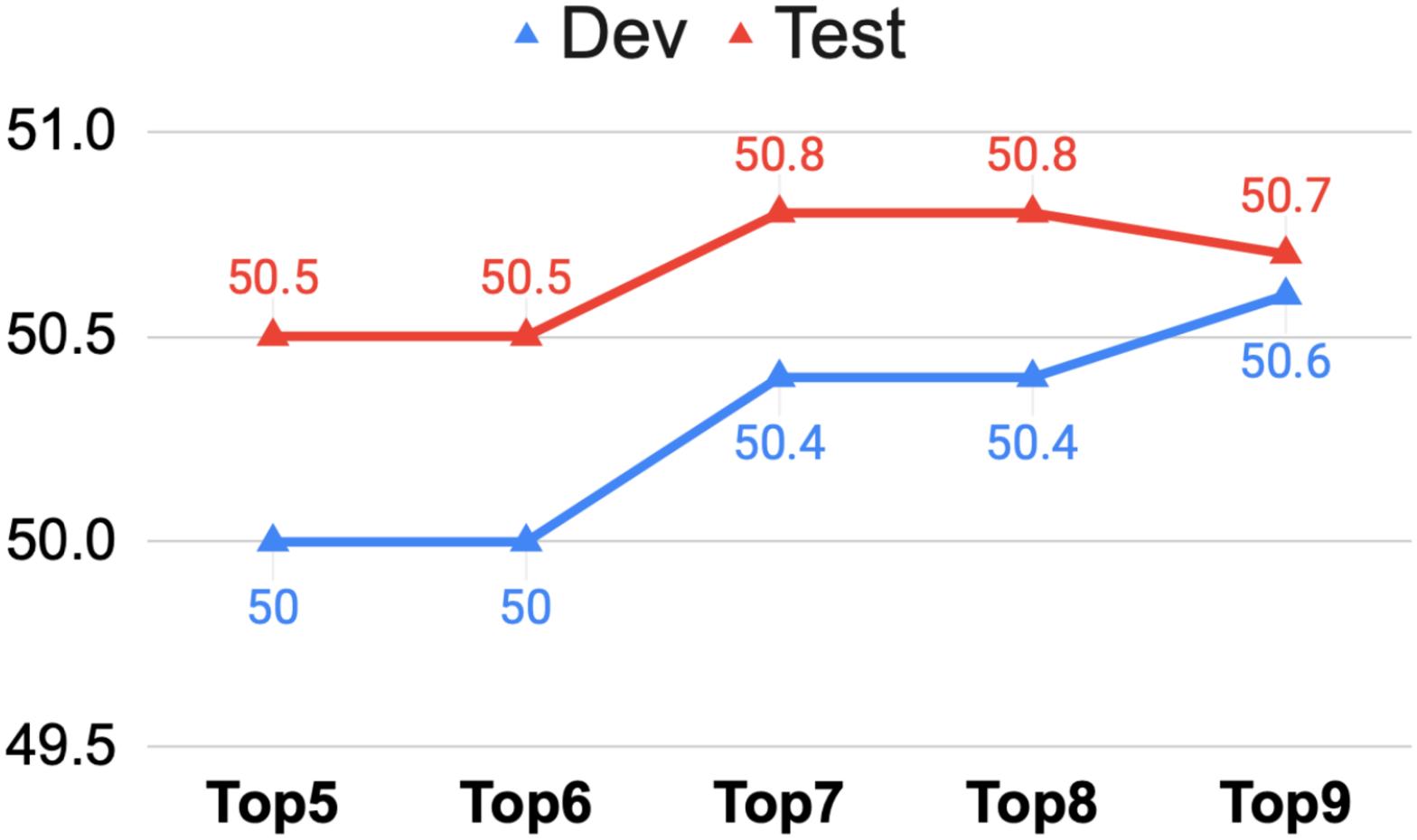}
    \caption{Dev v.s. Test sets performance ($F_1$ score) after relaxing the threshold of triplet count for selecting constraints. All numbers are percentages.}
    \label{fig:tbd-relaxed-constraints}
    \vspace{-0.5cm}
\end{figure}

In Sec.~\ref{sec:imp}, we use a 3\% threshold when selecting candidate constraints. In this section, we show the impact of relaxing this threshold on {\tbd}. Table~\ref{tab:tbd-triplets} shows three constraints that miss the 3\% bar by 0.1-0.3\%. In Figure~\ref{fig:tbd-relaxed-constraints}, we show F$_1$ scores on the development and test sets by including these constraints. Recall that only constraints that do not hurt development F$_1$ score are used. Therefore, \textbf{Top5} and \textbf{Top6} on the chart both correspond to the results in Table~\ref{tab:result}. \textbf{Top7} includes (\type{reporting}, \type{reporting}, \temprel{vague}), \textbf{Top8} includes (\type{actioin}, \type{reporting}, \temprel{vague}), and \textbf{Top9} includes (\type{reporting}, \type{actioin}, \temprel{vague}).

We observe that F$_1$ score continues to improve over the development set, but on the test set, F$_1$ score eventually falls. This appears to support our hypothesis that when the triplet count is small, the ratio calculated based on that count is not so reliable as the ratio could vary drastically between development and test sets. Optimizing over the development set can be an over-correction for the test set, and hence results in a performance drop.

\subsection{Event Type Prediction}
As described in Sec~\ref{sec:baselines}, to ensure fair comparison with the previous SOTA system \cite{han-etal-2019-joint}, our baseline model for {\tbd} does not predict event types. That is, when counting the triplet $(P^m, P^n, \hat{r})$, we assume there is an oracle model that provides event types $P^m, P^n$ for the predicted relation $\hat{r}$. One could potentially extend our work by training a similar multi-task learning model to predict both types and relations as our model does for the {\ib} dataset. We leave this as a future research direction.
\section{Related Work}
\label{sec:related}
\paragraph{News Domain.}
 Early work on \textbf{temporal relation extraction} use local pair-wise classification with hand-engineered features~\cite{Mani:2006:MLT:1220175.1220270, Verhagen:2007:STT:1621474.1621488, Chambers:2007:CTR:1557769.1557820,Verhagen:2008:TPT:1599288.1599300}. Later efforts, such as ClearTK \citep{S13-2002}, UTTime \citep{laokulrat-EtAl:2013:SemEval-2013}, NavyTime \cite{chambers:2013:SemEval-2013}, and CAEVO \citep{ChambersTBS2014}, improve earlier work with better linguistic and syntactic rules.
\citet{yoshikawa2009jointly,NingFeRo17,leeuwenberg2017structured} explore structured learning for this task, and
more recently, neural methods have also been shown effective~\cite{tourille2017neural, cheng2017classifying, meng2017temporal, meng2018context}. \citet{ning-etal-2018-cogcomptime} and \citet{han-etal-2019-joint} are the most recent work leveraging neural network and pre-trained language models to build an end-to-end system. Our work differs from these prior work in that we build a structured neural model with \textit{distributional constraints} that combines both the benefits of both deep learning and \textit{domain knowledge}.

\paragraph{Clinical Domain.}
The 2012 i2b2 Challenge (\cite{rumshisky-etal-2013-eval}) is one of the earliest efforts to advance event temporal relation extraction of clinical data. The challenge hosted three tasks on event (and event property) classification, temporal relation extraction, and the end-to-end track. Following this early effort, a series of clinical event temporal relation challenges were created in the following years (\cite{bethard-etal-2015-semeval, bethard-etal-2016-semeval, bethard-etal-2017-semeval}). However, data in these challenges are relatively hard to acquire, and therefore they are not used in this paper. As in the news data, traditional machine learning approaches~\cite{lee-etal-2016-uthealth,chikka-2016-cde,xu2013end,tang2013hybrid,savova2010mayo} that tackle the end-to-end event and temporal relation extraction problem require time-consuming feature engineering such as collecting lexical and syntax features. Some recent work~\cite{dligach-etal-2017-neural,leeuwenberg2017structured,galvan-etal-2018-investigating} apply neural network-based methods to model the temporal relations, but are not capable of incorporating prior knowledge about clinical events and temporal relations as proposed by our framework.
\section{Conclusion}
In conclusion, we propose a general framework that augments deep neural networks with \textit{distributional constraints} constructed using probabilistic \textit{domain knowledge}. We apply it in the setting of end-to-end temporal relation extraction task with event-type and relation constraints and show that the MAP inference with \textit{distributional constraints} can significantly improve the final results.

We plan to apply the proposed framework on various event reasoning tasks and construct novel \textit{distributional constraints} that could leverage \textit{domain knowledge} beyond corpus statistics, such as the larger unlabeled data and rich information contained in knowledge bases.

\section*{Acknowledgments}
This work is supported by the Intelligence Advanced Research Projects Activity (IARPA), via Contract No. 2019-19051600007 and the US Defense Advanced Research Projects Agency (DARPA), via Contract W911NF-15-1-0543. 
The views expressed are those of the authors and do not reflect the Department of Defense's official policy or position or the U.S. Government. 

\bibliography{emnlp2020}
\bibliographystyle{acl_natbib}
\appendix

\section*{Appendix}
\label{sec:appendix}

\section{Hyper-parameters}
\begin{table}[h]
\small
\centering
\begin{tabular}{lcc}
\toprule
 & \tbd  & \ib  \\ \hline \hline
$c_{\mathcal{E}}$ & - & 1.0 \\
$w_{\mathcal{E}_{pos}}$ & - & 5.0 \\ 
lr & -  & $2e^{-5}$ \\ 
$\alpha$ & 5.0 & 5.0 \\
$\theta$ & 0.05 & 0.02 \\ 
$\gamma$ & 0.7   & 0.8 \\ 
\bottomrule
\end{tabular}
\caption{Hyper-parameters chosen using development data. For \tbd, end-to-end baseline model is provided by the \citet{han-etal-2019-joint}, so we do not train it from scratch.}
\label{tab:hyper}
\end{table}

\section{Data Summary}
 \begin{table}[h]
 \small
 	\centering
 	\begin{tabular}{l|c|c} \hline
 	& \tbd & \ib \\ \hline\hline
    \multicolumn{3}{c}{\textbf{\# of Documents}} \\ \hline
 	Train & 22 &  190\\
 	Dev  &  5 &  -\\
    Test & 9 &  120\\\hline\hline
    \multicolumn{3}{c}{\textbf{\# of Pairs}} \\ \hline
 	Train & 4032 &  11253\\
 	Dev & 629 & - \\
 	Test & 1427 & 8794 \\\hline
 	\end{tabular}
   	\caption{Data overview. Note that we exclude event pairs whose sentence distance longer than 3 in {\ib}, and there are 6 times more missing relations than the gold annotated ones in, which explains why number of pairs per documents are smaller in {\ib} than in {\tbd}.}
   	\label{tab:data}
 \end{table}

\section{{\ib} Results}
\label{sec:i2b2-results}
We show the breakdown performance and contributions of individual constraints for {\ib} in Table~\ref{tab:i2b2-breakdown} and Table~\ref{tab:i2b2-abl} respectively.
\begin{table}[h]
    \small 
 	\centering
 	\setlength{\tabcolsep}{0.5em}
 	\begin{tabular}{|@{ }c@{}|c@{ }|c@{ }|c@{ }|c@{ }|c@{ }|c@{ }|} \hline
 	&\multicolumn{3}{|@{ }c@{ }|}{\textbf{End-to-end Baseline}}
 	& \multicolumn{3}{|@{ }c@{ }|}{\textbf{End-to-end Inference}}\\ \cline{2-7}
 	& \textbf{P}& \textbf{R} & \textbf{F$_1$}&\textbf{P}& \textbf{R} & \textbf{F$_1$}\\ \hline
 	 \textbf{B}& 82.1& 60.6& 69.7 & 80.9& 65.3& 72.2\\
 	 \textbf{A} & 69.9 & 59.9 & 64.5 & 67.8& 62.8& 65.2\\
 	 \textbf{O}&81.3 & 81.5 & 81.4 & 83.6 & 80.2 & 81.9\\ \hline
 	 \textbf{TempEval}&73.3&  79.9& 76.5 & 74.0 & 80.8 & 77.3\\ \hline
 	\end{tabular}
   	\caption{Model performance breakdown for {\ib}. \small{\temprel{Before} (\textbf{B}), \temprel{After} (\textbf{A}), \temprel{overlap} (\textbf{O}).}}
   	\label{tab:i2b2-breakdown}
 \end{table}
 
 \begin{table}[h]
\centering
\small
\begin{tabular}{@{ }l@{ }c@{ }c@{ }}
\toprule
Constraint Triplets & Prob. Gap  & F$_1$  \\ \hline \hline
\type{occur.}, \type{problem}, \temprel{overlap} & -0.17 & +0.19\% \\
\type{occur.}, \type{treatment}, \temprel{overlap} & -0.16 & +0.24\% \\ 
\type{treatment}, \type{occur.}, \temprel{overlap} & -0.11   & +0.22\% \\ 
\type{treatment}, \type{problem}, \temprel{overlap} & -0.14  & +0.12\% \\
\hline\hline
Combined F1 Improvement & & 0.77\% \\
\bottomrule
\end{tabular}
\caption{{\ib} ablation study: gap shrinkage of predicted probability and F$_1$ contribution per constraint.}
\label{tab:i2b2-abl}
\vspace{-0.3cm}
\end{table}

\section{Reproducibility List}
\begin{itemize}
    \item Data and code used for {\tbd} can be found in project code base. However, due to user confidentiality agreement, we are not able to provide data and and data analysis code for {\ib}. Modeling code will be added to the project code base upon obtaining permission from the data owner. 
    \item We use BERT-base-uncased and Roberta-large models implemented in Huggingface transformers. Additional parameters (such as LSTM and MLP) are negligible compared to those used in the pre-trained LMs;
    \item ILP is solved by an off-the-shelf solver provided by Gurobi optimizer;
    \item Range of grid-search. $c_{\mathcal{E}}$: (1.0, 2.0); $w_{\mathcal{E}_{pos}}$: (1.0, 2.0, 5.0, 10.0); lr: ($1e^{-5}$, $2e^{-5}$, $5e^{-5}$), $\alpha$: (1.0, 2.0, 5.0, 10.0); $\theta$: (0.2, 0.3, 0.5); $\gamma$: (0.7, 0.8, 0.9).
\end{itemize}

\end{document}